\documentclass[pdflatex,sn-mathphys-num]{sn-jnl}

\usepackage{graphicx}%
\usepackage{subcaption}
\usepackage{multirow}%
\usepackage{amsmath,amssymb,amsfonts}%
\usepackage{amsthm}%
\usepackage{mathrsfs}%
\usepackage[title]{appendix}%
\usepackage{xcolor}%
\usepackage{textcomp}%
\usepackage{manyfoot}%
\usepackage{booktabs}%
\usepackage{algorithm}%
\usepackage{algpseudocode}%
\usepackage{algorithmicx}%
\usepackage{listings}%

\usepackage{tabularx}
\usepackage{rotating}
\usepackage{threeparttable}

\theoremstyle{thmstyleone}%
\theoremstyle{thmstyletwo}%

\theoremstyle{thmstylethree}%

\begin{document}

\title[Article Title]{Dual-Domain Sparse-View 4D CBCT Reconstruction with Spatial-Temporal Consistency Learning}

\author[1]{\sur{Qi Lai}}\email{q.lai@siat.ac.cn}

\author[1,2]{\sur{YuTong He}}\email{heyt3@mail.sustech.edu.cn}

 \abstract{
Four-dimensional cone-beam computed tomography (4D CBCT) provides phase-resolved volumetric images for image-guided radiotherapy. Yet good reconstruction quality usually needs many projections. This increases scan time and radiation dose. We propose STCNet, a sparse-view 4D CBCT reconstruction framework that uses spatial-temporal consistency learning. STCNet reconstructs high-quality images from only 16 projections in each respiratory phase. STCNet integrates physics-driven positional encoding to enforce geometric consistency in the projection domain and introduces temporal encoding in latent space to capture respiratory motion dynamics across phases to capture dynamic respiratory motion across phases. A dual-domain collaborative strategy jointly optimizes both projection and image domains, enhancing structural fidelity and temporal coherence. We tested STCNet on the TCIA-LUNG dataset and a CIRS thorax phantom dataset. STCNet performed better than conventional methods and recent deep learning baselines in image quality, artifact reduction, and structural preservation. On TCIA-LUNG, STCNet reached PSNR values of 33.311, 32.695, and 32.658 dB at phases 2, 5, and 8. The corresponding SSIM values were 0.845, 0.823, and 0.825. On CIRS, STCNet achieved PSNR values up to 36.497 dB and SSIM values up to 0.877. These results support STCNet as a promising framework for low-dose, phase-resolved 4D CBCT reconstruction in image-guided radiotherapy.
}




\maketitle

\section{Introduction}
\label{sec:Introduction}

Image-guided radiation therapy (IGRT) has become integral to contemporary radiotherapy by enabling treatment delivery to be aligned with patient anatomy at the time of irradiation \cite{franzone2016image,cuijpers2017analysis,caillet2017igrt,johnson2018residual}. This role is especially critical for thoracic malignancies, where respiratory motion can introduce substantial uncertainty in tumor position during treatment and thereby compromise targeting accuracy \cite{caillet2017igrt,johnson2018residual,roman2012interfractional,r14}. In this setting, robust on-board imaging is essential not only for setup verification, but also for motion-aware assessment of the tumor and surrounding normal tissues \cite{franzone2016image,cuijpers2017analysis,caillet2017igrt,johnson2018residual}.
Cone-beam computed tomography (CBCT) is widely used for volumetric image guidance on the treatment unit. \cite{franzone2016image,cuijpers2017analysis,caillet2017igrt,johnson2018residual,roman2012interfractional} However, conventional three-dimensional CBCT has well-recognized limitations in the thorax because respiratory motion is averaged into a single reconstructed volume, often causing motion blur, streak artifacts, and loss of boundary definition \cite{caillet2017igrt,r14,ou2024pnmc,qi2011extraction}. These degradations can reduce confidence in target localization and limit the reliability of image guidance for moving lesions \cite{caillet2017igrt,johnson2018residual,roman2012interfractional,r14}. 

Four-dimensional CBCT (4D CBCT) was developed to overcome this limitation by sorting projection data into respiratory-correlated bins and reconstructing phase-resolved image volumes \cite{ou2024pnmc,zhi2021cycn}. By preserving temporal information across the breathing cycle, 4D CBCT can improve visualization of moving targets and provide a more informative representation of respiratory motion than conventional CBCT \cite{qi2011extraction,brehm2013artifact,wang2013simultaneous,jia2012four,zhang2017high}. This capability is particularly relevant for radiotherapy workflows that depend on accurate motion characterization, including target verification, treatment adaptation, and motion management \cite{caillet2017igrt,brehm2013artifact,zhang2017high}. 
Despite its clinical appeal, routine use of 4D CBCT remains constrained by image quality limitations. Because the acquired projections are divided across respiratory phases, each phase is reconstructed from only a small subset of views, making 4D CBCT inherently a sparse-view problem \cite{ou2024pnmc,zhang2017high,mascolo2018fast}. The resulting undersampling can produce severe streaking artifacts, structural distortion, and unstable intensity representation, particularly when the number of projections per phase is further reduced to limit imaging dose \cite{brehm2013artifact,jia2012four,zhang2017high,mascolo2018fast}. These limitations directly affect the clinical interpretability of phase-resolved images.
Dose efficiency is therefore a central challenge in 4D CBCT. In practice, obtaining acceptable image quality often requires many projections acquired over multiple breathing cycles, increasing scan time and imaging dose \cite{yang2022four}. This is undesirable in radiotherapy, where repeated on-board imaging may be needed throughout a treatment course \cite{franzone2016image,caillet2017igrt,johnson2018residual}.  Methods that enable high-quality reconstruction from substantially fewer projections could therefore improve the practicality of low-dose 4D CBCT for routine clinical guidance \cite{chan2023attention,zhang2024xtransct}.
A broad range of strategies has been proposed to address sparse-view 4D CBCT reconstruction. Motion-compensated methods exploit inter-phase deformation to reduce respiratory artifacts and improve reconstruction fidelity \cite{ou2024pnmc,brehm2013artifact}. Regularized iterative approaches, including temporal nonlocal and motion-compensated formulations, can further suppress noise and undersampling artifacts \cite{wang2013simultaneous,jia2012four,mascolo2018fast}. However, these approaches often depend on accurate motion estimation, may be computationally intensive, and can oversmooth fine structures when the acquisition is extremely sparse.
Deep learning has recently emerged as a powerful alternative for sparse-view CT and CBCT reconstruction \cite{zhang2024xtransct,dong20232d,li2019learning,8452958,zhang2018sparse,cao2022swin,xie2018artifact}. Prior studies have explored image-domain artifact reduction \cite{chan2023attention,xie2018artifact,han2018framing},  sinogram- or projection-domain synthesis \cite{li2019learning,8452958}, ultra-sparse-view volumetric reconstruction \cite{zhang2018sparse,zhang2024xgenrecon}, and generative modeling frameworks based on diffusion or flow models \cite{xia2022patch,liu2022learning}.  Deep learning has also been used to derive four-dimensional CBCT information from highly limited acquisition data, highlighting the potential of data-driven reconstruction for low-dose imaging \cite{yang2022four}. Nevertheless, many existing methods primarily address either data completion or image enhancement in isolation, without fully exploiting the coupled geometric structure of projection acquisition and the temporal continuity of respiratory motion that define the 4D CBCT problem \cite{li2019learning,heylen20194d,zhang2024xgenrecon}. 
For 4D CBCT, these two forms of prior information are fundamentally complementary. Projection measurements are governed by fixed acquisition geometry, whereas reconstructed anatomy evolves in a temporally correlated manner across respiratory phases \cite{xie2018artifact,wang2013simultaneous,jia2012four,zhang2017high}. A reconstruction framework that jointly models both relationships may therefore recover missing information more faithfully than approaches operating in only a single domain or without explicit temporal consistency \cite{yang2022four,zhang2024xtransct,zhang2024xgenrecon}. 
In this work, we propose STCNet, a sparse-view 4D CBCT reconstruction framework based on spatial-temporal consistency learning. STCNet is designed for low-dose phase-resolved imaging using only 16 projections per respiratory phase. The framework combines projection-domain completion with image-domain refinement after Feldkamp-Davis-Kress reconstruction. In the projection domain, physics-aware positional encoding is used to represent angular acquisition geometry, while temporal encoding is introduced to capture respiratory dynamics across phases. The completed projections are then reconstructed and further refined to improve structural fidelity and suppress residual sparse-view artifacts.

\begin{figure*}[h]
    \centering
    \includegraphics[width=\textwidth]{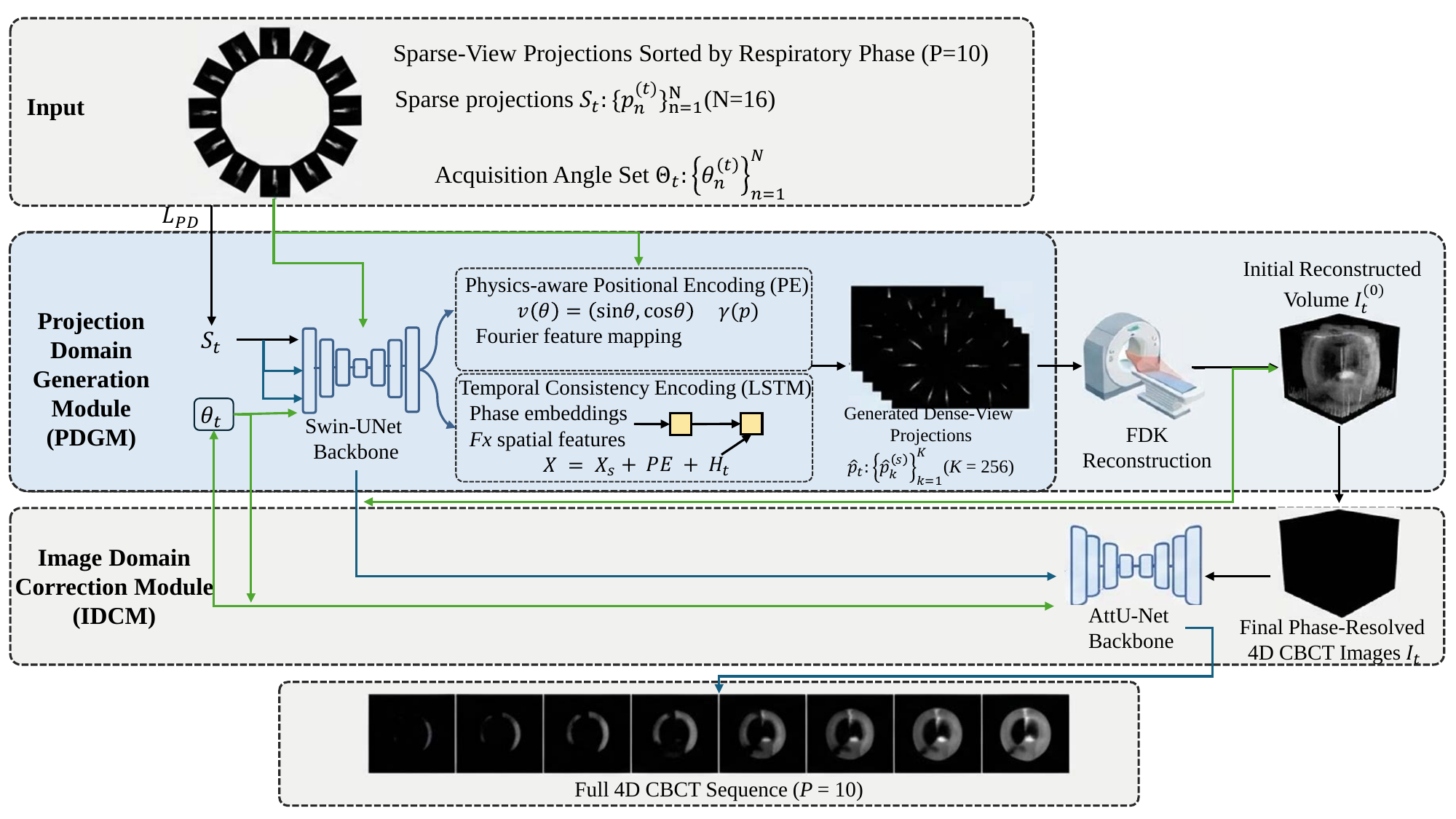}
    \caption{Overview of the proposed STCNet framework for dual-domain sparse-view 4D CBCT reconstruction with spatial-temporal consistency learning. Sparse-view projections are sorted into $P=10$ respiratory phases, with $N=16$ projections per phase. The projection-domain generation module (PDGM) incorporates physics-aware positional encoding of acquisition angles and LSTM-based temporal consistency encoding to generate $K=256$ dense-view projections per phase. FDK reconstruction converts the generated projections into initial phase-resolved volumes, which are subsequently refined by the image-domain correction module (IDCM) to produce the final 4D CBCT sequence.}
    \label{fig:stcnet_overview}
\end{figure*}

We evaluate STCNet using the TCIA-LUNG dataset \cite{r15,r16}. Its performance is compared with conventional reconstruction approaches and representative deep learning baselines under an ultra-sparse acquisition setting. Our aim is to determine whether explicit modeling of geometric and temporal consistency can improve image quality and anatomical fidelity, and thereby support a more clinically feasible pathway toward low-dose 4D CBCT for image-guided radiotherapy.

\section{Results}
\label{sec2}
\subsection{Dataset Definition}
\label{ssec:dataset}

Two datasets were used to evaluate the proposed STCNet: CIRS thorax phantom dataset for experimental validation and the publicly available TCIA-LUNG dataset \cite{roman2012interfractional,r14,r15,r16} for evaluating generalization performance. The CIRS dataset was acquired in our laboratory and is a private dataset, providing experimentally measured CBCT projections under controlled respiratory motion. The TCIA-LUNG dataset was used to assess the applicability of the proposed method beyond the controlled phantom setting. The acquisition and reconstruction parameters are summarized in Table~\ref{tab:acq_params}.

\begin{table}[!h]
  \centering
  \caption{Acquisition parameters for the TCIA-LUNG and CIRS phantom datasets}
  \label{tab:acq_params}
  \begin{tabular}{lcc}
    \midrule
     \toprule[1pt]
    Parameter & TCIA-LUNG \cite{roman2012interfractional,r14,r15,r16} & CIRS phantom \\
    \midrule
    DSD/DSO (mm) & 1500/1000 & 1800/1000 \\
    Pixel size (mm$^2$) & 0.95 $\times$ 0.95 & 0.95 $\times$ 0.95 \\
    Detector size & 1024 $\times$ 1024 & 1024 $\times$ 1024 \\
    Angle range & 0$^\circ$--360$^\circ$ & 0$^\circ$--360$^\circ$ \\
    Voxel size (mm$^3$) & 0.98 $\times$ 0.98 $\times$ 1 & 0.98 $\times$ 0.98 $\times$ 1 \\
    Dimension & 512 $\times$ 512 $\times$ N$_z$ & 512 $\times$ 512 $\times$ N$_z$ \\
    \midrule
    \bottomrule[1pt]
  \end{tabular}
  \begin{tablenotes}
    \footnotesize
    \item[*]\textbf{Remark}: : N$_z$ denotes the number of slices, which varied across cases/datasets.
  \end{tablenotes}
\end{table}

\noindent \textbf{CIRS Thorax Phantom Dataset}
To evaluate the reconstruction performance of STCNet under controlled experimental conditions, a dynamic CIRS thorax phantom was scanned using a CBCT imaging system in our laboratory. The phantom was scanned along a circular acquisition trajectory and contained a spherical target serving as a tumor surrogate. During acquisition, the target underwent prescribed sinusoidal motion to simulate respiratory-induced tumor displacement.

A total of 2,560 projection images were acquired over a full $360^{\circ}$ rotation, with an angular interval of approximately $0.14^{\circ}$ between adjacent views. Projections were acquired continuously while the target followed the predefined respiratory trajectory. Each projection was assigned to a respiratory phase according to its position within the motion cycle. The acquired projections were retrospectively sorted into 10 respiratory phase bins, with projections corresponding to the same phase across different respiratory cycles grouped together. These phase-specific projection subsets were used for subsequent sparse-view reconstruction experiments.

The source-to-object distance (SOD) and source-to-detector distance (SDD) were 1,000~mm and 1,800~mm, respectively. The detector comprised $3072 \times 3072$ pixels, with a physical pixel size of $0.172 \times 0.172~\mathrm{mm}^{2}$. The X-ray tube voltage and current were set to 120~kVp and 0.4~mA, respectively, with an exposure time of \textit{[to be specified]}~ms per projection.

Before reconstruction, the acquired projections were corrected using the corresponding flat-field and dark-field images. The corrected projections were subsequently converted into line-integral measurements through a negative logarithmic transformation. The reconstructed volumes had a matrix size of \textit{[to be specified]} voxels and a voxel spacing of \textit{[to be specified]}~mm. Unless otherwise specified, all reconstruction methods evaluated on the CIRS dataset used the same acquisition geometry and preprocessing pipeline to ensure a consistent comparison.

The CIRS dataset was used for experimental validation of STCNet using physically acquired sparse-view CBCT projections. This controlled setting enabled evaluation of phase-resolved reconstruction quality and recovery of the moving target in the presence of respiratory motion.

\noindent \textbf{TCIA-LUNG Dataset}
The publicly available TCIA-LUNG dataset \cite{roman2012interfractional,r14,r15,r16} was used to evaluate the generalization performance of STCNet beyond the controlled phantom setting. In contrast to the CIRS phantom experiments, this dataset provided patient anatomical structures for assessing the applicability of the proposed method to more complex thoracic anatomy.

\subsection{Comparison Methods}

We compared STCNet with both conventional reconstruction methods and representative deep learning baselines. The conventional methods included FDK, simultaneous algebraic reconstruction technique (SART) \cite{andersen1984simultaneous}, and SART-TV / OSCTV-based sparse-view reconstruction \cite{mascolo2018fast}. The deep learning baselines included N-Net, CyCN-Net \cite{zhi2021cycn}, and AttUNet \cite{chan2023attention} . We also evaluated the projection-domain module alone (PDGM) to assess the contribution of dual-domain refinement.

\subsection{Evaluation Metrics}

We evaluated reconstruction quality using mean absolute error (MAE) \cite{chai2014root}, peak signal-to-noise ratio (PSNR) \cite{poobathy2014edge}, and structural similarity index (SSIM) \cite{chen2011fast}. MAE measures the absolute intensity difference between the reconstructed image and the reference image, with lower values indicating better reconstruction accuracy.  PSNR quantifies fidelity in terms of signal-to-error ratio, with higher values indicating better image quality.SSIM measures structural similarity and perceptual consistency between images, with higher values indicating better structural preservation.

\subsection{Implement Details}

All experiments were implemented in Python using the PyTorch framework. Training and validation were conducted on a workstation equipped with an Intel CPU, 128 GB RAM, and an NVIDIA A6000 GPU with 48 GB memory.

For TCIA-LUNG, each phase volume was forward projected into 256-view dense projection data, and 16 uniformly sampled views were selected as sparse-view input. The PDGM and IDCM were trained as the two sequential components of the overall framework. During inference, the trained STCNet model took sparse-view phase-resolved projection data as input and generated the final reconstructed 4D CBCT volumes.

\subsection{Quantitative Comparison}

Table~\ref{tab:main_results} presents the quantitative results on axial slices from the CIRS thorax phantom and TCIA-LUNG datasets at respiratory phases 2, 5, and 8.


\renewcommand{\arraystretch}{1.3}
\begin{table*}[!h]
\caption{ Quantitative comparison of reconstruction performance on axial slices from the CIRS Phantom and TCIA-LUNG datasets. \label{tab:main_results}}
\centering
\footnotesize
\begin{threeparttable}
\begin{tabular}{lccccccccc}
\midrule
\toprule[1pt]

\multirow{3}{*}{Methods} & \multicolumn{9}{c}{CIRS Phantom} \\ \cmidrule(lr){2-10} 
&\multicolumn{3}{c}{Phase2} & \multicolumn{3}{c}{Phase5} &\multicolumn{3}{c}{Phase8}\\\cmidrule(lr){2-10}
& MAE$\downarrow$ &PSNR$\uparrow$ & SSIM$\uparrow$     & MAE$\downarrow$ &PSNR$\uparrow$ & SSIM$\uparrow$  & MAE$\downarrow$ &PSNR$\uparrow$ & SSIM$\uparrow$   \\
\hline
FDK \cite{feldkamp1984practical} &0.196&22.316&0.324 &0.194&22.289&0.349
&0.189&22.326&0.335 \\
SART \cite{andersen1984simultaneous} &0.129
&25.893&0.349&0.123&25.719&0.346
&0.124&25.892&0.349 \\
SART-TV \cite{mascolo2018fast} &0.113&26.735&0.457& 0.117&26.741&0.461
&0.121&26.658&0.449 \\
N-Net \cite{zhi2021cycn} &0.071 &20.322 &0.506 &0.071 &20.313 &0.505
&0.072 &20.126 &0.499 \\
CycN-Net \cite{zhi2021cycn} &0.102 &23.338 &0.423 &0.102 &23.239 &0.422 
&0.103 &23.160 &0.421 \\
AttUnet\cite{chan2023attention}  &0.068 &24.493&0.261&0.066&24.358&0.265
&0.062&24.374&0.258 \\
PGDM &\underline{0.063} &\underline{30.513} &\underline{0.711} &\underline{0.061} &\underline{30.408} &\underline{0.717} &\underline{0.065} &\underline{30.501} &\underline{0.710} \\
STCNet &\textbf{0.018}&\textbf{36.497}&\textbf{0.871}&\textbf{0.017}&\textbf{36.392}&\textbf{0.874}
&\textbf{0.018} &\textbf{36.261}&\textbf{0.877} \\

\midrule
\bottomrule[1pt]

\multirow{3}{*}{Methods} & \multicolumn{9}{c}{TCIA-LUNG} \\ \cmidrule(lr){2-10} 
&\multicolumn{3}{c}{Phase2} & \multicolumn{3}{c}{Phase5} &\multicolumn{3}{c}{Phase8}\\\cmidrule(lr){2-10}
& MAE$\downarrow$ &PSNR$\uparrow$ & SSIM$\uparrow$     & MAE$\downarrow$ &PSNR$\uparrow$ & SSIM$\uparrow$  & MAE$\downarrow$ &PSNR$\uparrow$ & SSIM$\uparrow$   \\
\hline
FDK \cite{feldkamp1984practical} &0.180&21.585&0.309&0.174&22.077&0.317
&0.182&21.687&0.308 \\
SART \cite{andersen1984simultaneous} &0.073&22.573&0.536&0.076&22.360&0.530
&0.068&22.177&0.528 \\
SART-TV \cite{mascolo2018fast} &0.071&23.453&0.540& 0.064&23.377&0.538
&0.067&23.264&0.537 \\
N-Net \cite{zhi2021cycn} &0.136 &22.025 &0.473 &0.129 &22.137 &0.468
&0.133 &22.021 &0.471 \\
CycN-Net \cite{zhi2021cycn} &0.113 &22.483 &0.564 &0.109 &22.514 &0.557
&0.107 &22.488 &0.561 \\
AttUnet \cite{chan2023attention}  &0.174&23.631&0.293&0.171&23.797&0.274
&0.168&24.124&0.292 \\
PGDM & \underline{0.059} & \underline{30.332} & \underline{0.713} & \underline{0.061} & \underline{29.744} & \underline{0.715} & \underline{0.058} & \underline{29.738} & \underline{0.713} \\
STCNet &\textbf{0.035}&\textbf{33.311}&\textbf{0.845}&\textbf{0.032}&\textbf{32.695}&\textbf{0.823}
&\textbf{0.039} &\textbf{32.658}&\textbf{0.825} \\

\midrule
\bottomrule[1pt]
\end{tabular}

 \begin{tablenotes}
    \footnotesize
    \item[*]\textbf{Remark}: the \textbf{Bold} and \underline{underlined} indicate this metric's best and second-best performance, respectively. $\uparrow$($\downarrow$) indicates that the larger (smaller) the value, the better the performance.
      \end{tablenotes}
    \end{threeparttable}
\end{table*}

STCNet achieved the lowest MAE and the highest PSNR and SSIM at all reported phases on both datasets (Table~\ref{tab:main_results}). On the CIRS thorax phantom dataset, SART-TV improved PSNR from 22.289--22.326~dB with FDK~\cite{feldkamp1984practical} to 26.658--26.741~dB, whereas the deep-learning baselines showed mixed performance across metrics. PDGM ranked second for all three metrics at each evaluated phase. STCNet further increased PSNR to 36.497, 36.392, and 36.261~dB at phases 2, 5, and 8, respectively, with corresponding SSIM values of 0.871, 0.874, and 0.877 and MAE values of 0.018, 0.017, and 0.018. Its PSNR gains of 5.760--5.984~dB over PDGM support the additional benefit of image-domain correction in the phantom experiments.

On the TCIA-LUNG dataset, SART and SART-TV improved all three metrics relative to FDK, while the deep-learning baselines showed mixed performance across metrics. PDGM ranked second at all evaluated phases, achieving PSNR values of 29.738--30.332~dB and SSIM values of 0.713--0.715. STCNet further increased PSNR to 33.311, 32.695, and 32.658~dB at phases 2, 5, and 8, respectively, with corresponding SSIM values of 0.845, 0.823, and 0.825 and MAE values of 0.035, 0.032, and 0.039. The PSNR gains of 2.920--2.979~dB over PDGM support the generalization performance of the dual-domain framework beyond the controlled phantom setting.

\begin{figure}[!h]
\centerline{\includegraphics[width=\columnwidth]{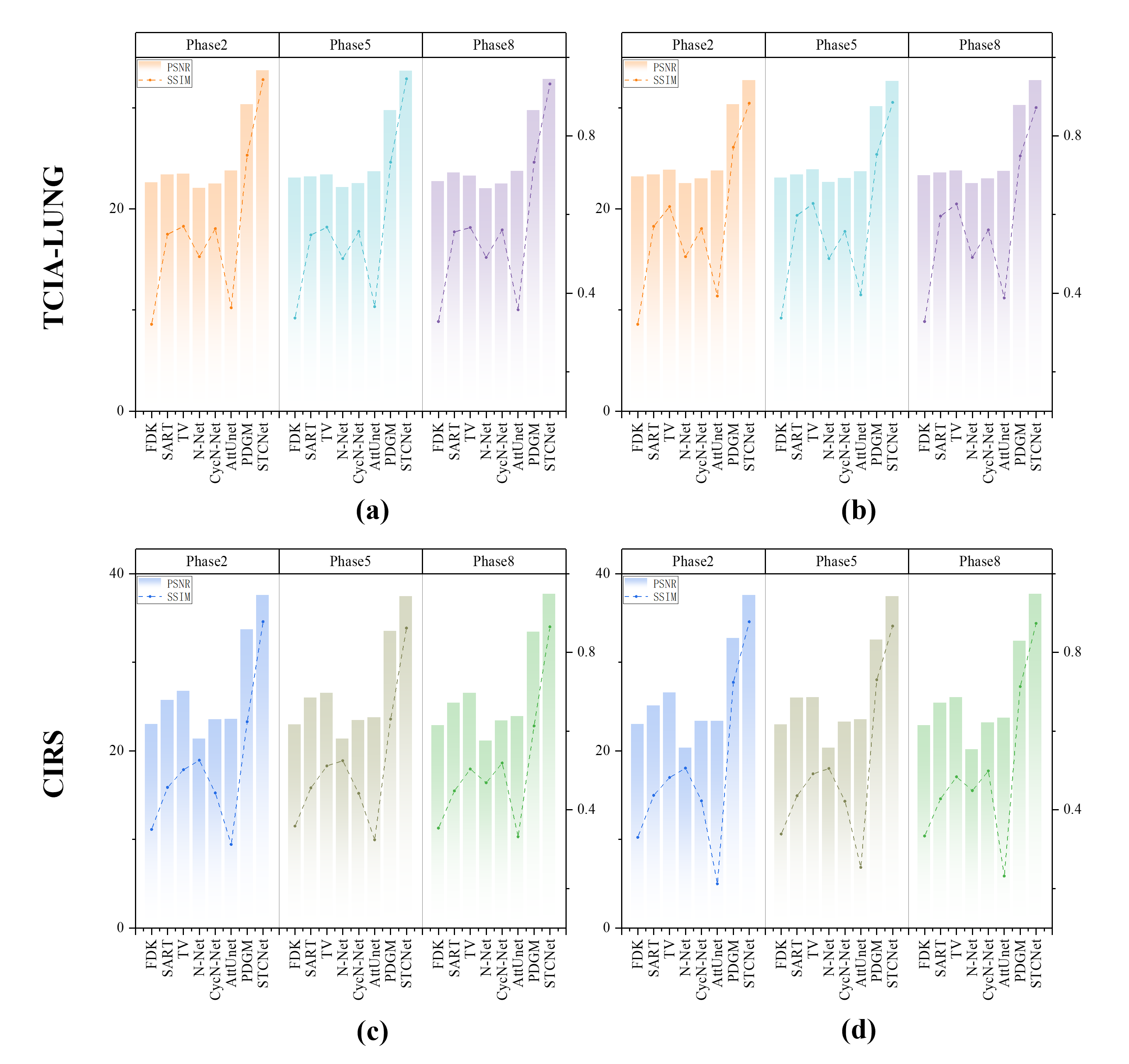}}
\caption{Quantitative comparison across planes and phases: The figure shows PSNR and SSIM results for representative phases in coronal and sagittal views. Panel (a) shows TCIA-LUNG coronal results. Panel (b) shows TCIA-LUNG sagittal results. Panel (c) shows CIRS coronal results. Panel (d) shows CIRS sagittal results.} 
\label{fig:zhexian}
\end{figure}

Figure~\ref{fig:zhexian} additionally compares PSNR and SSIM in the coronal and sagittal planes. The CIRS dataset was used for experimental validation, whereas the TCIA-LUNG dataset was used to assess generalization beyond the controlled phantom setting.
The coronal and sagittal comparisons in Fig.~\ref{fig:zhexian} show a similar trend: STCNet consistently achieved the highest PSNR and SSIM across the displayed phases on both datasets. Together with the axial-slice results, these findings demonstrate improved reconstruction fidelity and structural similarity across the evaluated anatomical planes and respiratory phases, supporting the complementary roles of projection-domain generation and image-domain correction.

\subsection {Reconstruction Quality}

\begin{figure*}[!h]
\centering
\includegraphics[width=5in]{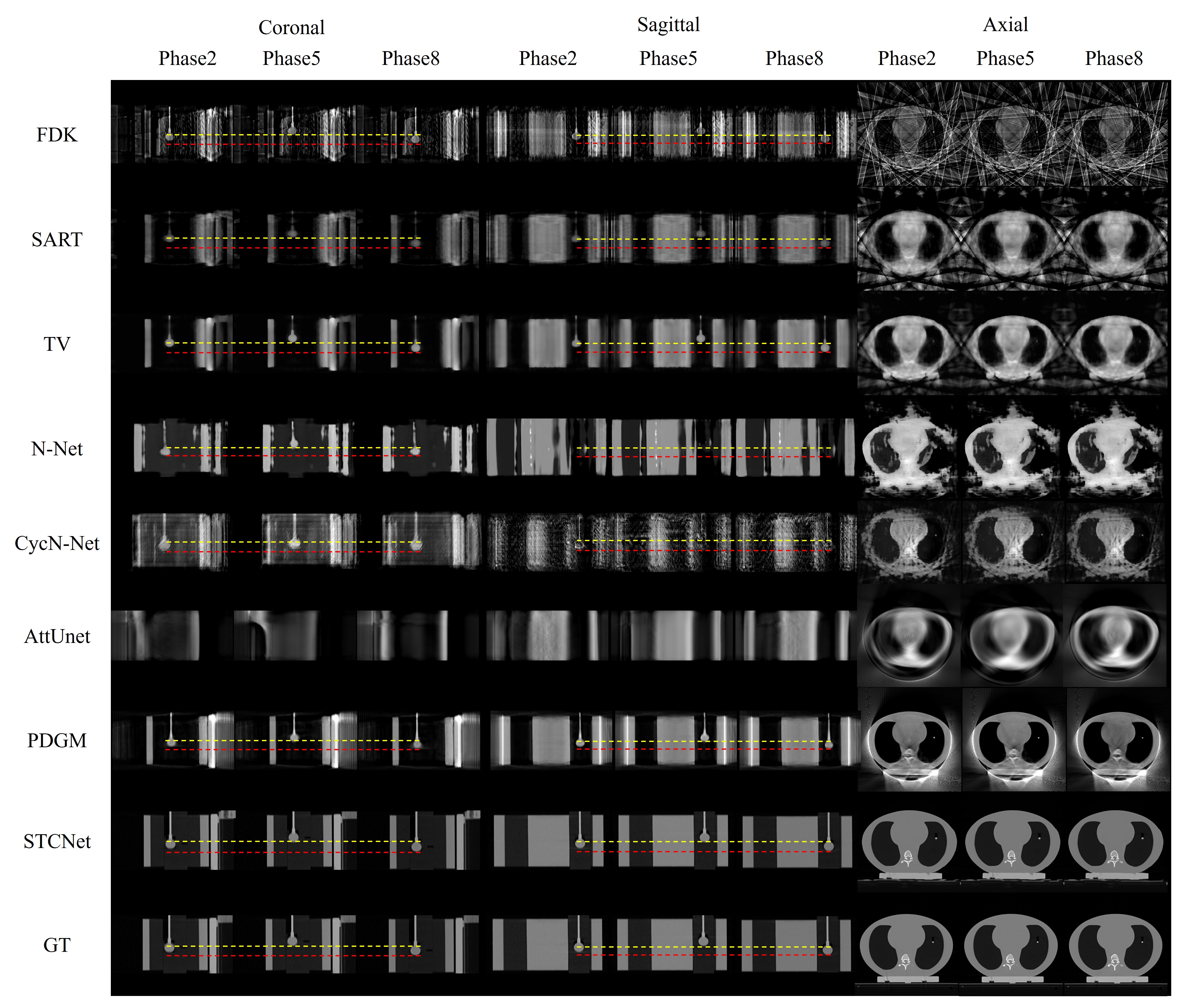}
\caption{Reconstruction results on the CIRS phantom dataset: The figure shows coronal, sagittal, and axial slices at phases 2, 5, and 8 for different methods. The red and yellow dashed lines mark tumor marker boundary positions across phases.}
\label{fig:main_resulets_CIRS}
\end{figure*}

We then tested the model on the CIRS thorax phantom dataset. Figure \ref{fig:main_resulets_CIRS} shows the results. 

The FDK \cite{feldkamp1984practical} again exhibited severe artifacts, particularly in the axial plane, which made tumor-marker boundaries difficult to identify. SART \cite{andersen1984simultaneous} and SART-TV \cite{mascolo2018fast} reduced some of these errors but still left visible artifacts and incomplete boundaries. The deep-learning baselines did not perform consistently: some methods improved smoothness, but they often missed fine structural details or distorted the marker boundaries.

STCNet provided the clearest reconstructions on this dataset. Marker boundaries were more accurate, structures appeared more continuous, and residual artifacts were fewer. These results suggest that STCNet can generalize beyond the training-like data setting.

\subsection {Generalization Ability}

\begin{figure*}[!h]
\centering
\includegraphics[width=5in]{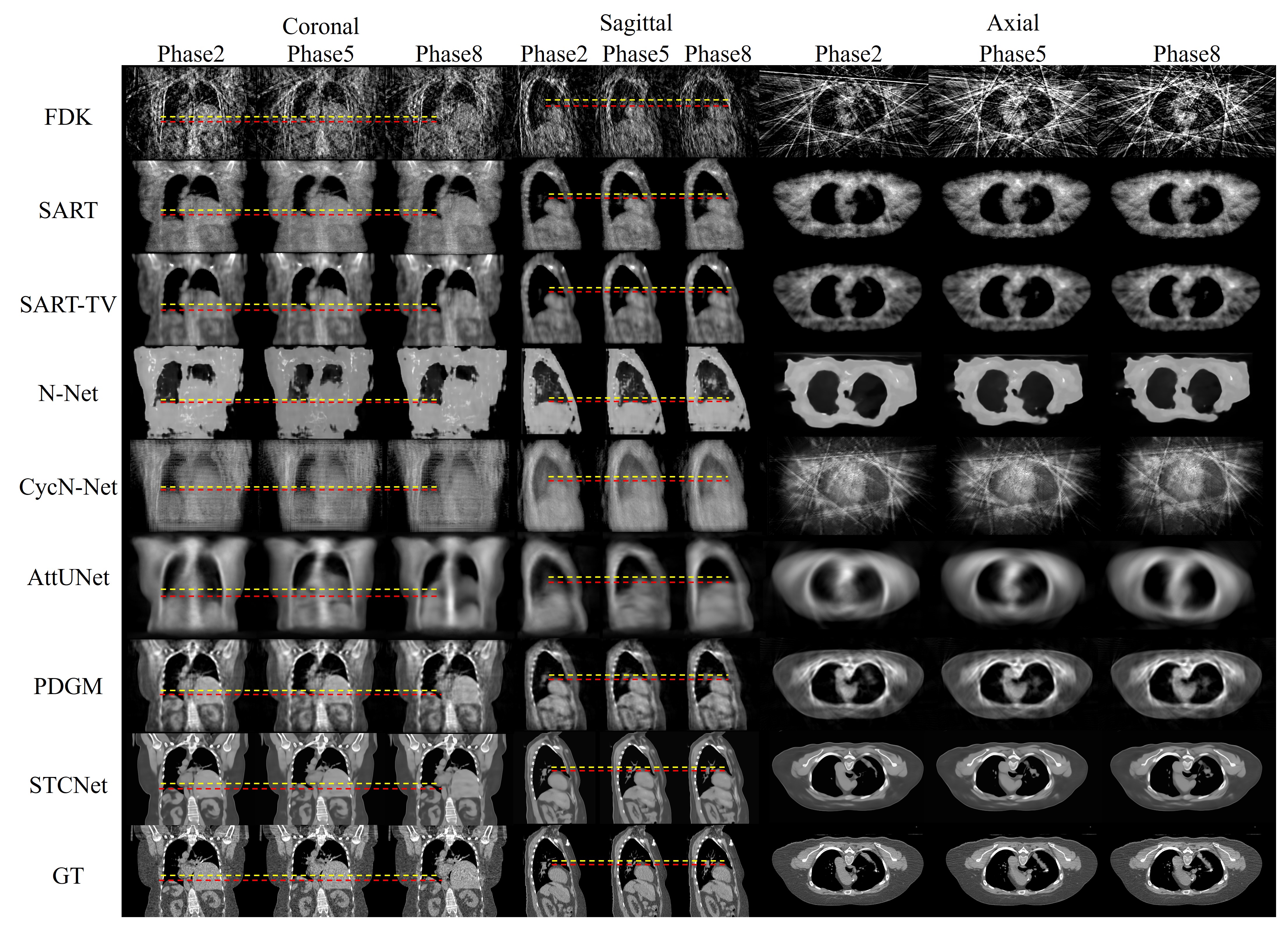}
\caption{Reconstruction results on the TCIA-LUNG dataset: The figure shows coronal, sagittal, and axial slices at phases 2, 5, and 8 for different methods. The red and yellow dashed lines mark the highest and lowest diaphragm positions and show the motion path across phases.}
\label{fig:main_resulets_TCIA}
\end{figure*}

We first tested STCNet on the TCIA-LUNG dataset. Figure \ref{fig:main_resulets_TCIA} shows coronal, sagittal, and axial results at phases 2, 5, and 8.

The FDK \cite{feldkamp1984practical} produced strong streak artifacts under the sparse-view setting. The ribs and spine looked blurred and broken. Soft tissue boundaries were also unclear. SART and SART-TV improved the overall appearance, but both methods still had visible noise and smoothing effects. SART \cite{andersen1984simultaneous} kept more structure than FDK, but it introduced noisy textures. SART-TV \cite{mascolo2018fast} reduced some noise, but it also blurred fine details.
The deep learning baselines showed mixed results. N-Net and CycN-Net \cite{zhi2021cycn} reduced some artifacts, but they did not recover fine structures well. AttUNet \cite{chan2023attention} produced smoother images, but it still caused anatomical distortion in some regions.

STCNet provided the best overall visual quality among the tested methods. The reconstructions showed fewer streak artifacts. The diaphragm region was more clearly delineated. Bone and soft tissue boundaries were more natural. The overall anatomical structure was also closer to the ground truth. PDGM alone already improved the image quality a lot. The full STCNet improved it further after image-domain correction.

\subsection{HU Error Map Analysis}

\begin{figure}[!h]
  \centering
  \includegraphics[width=\columnwidth]{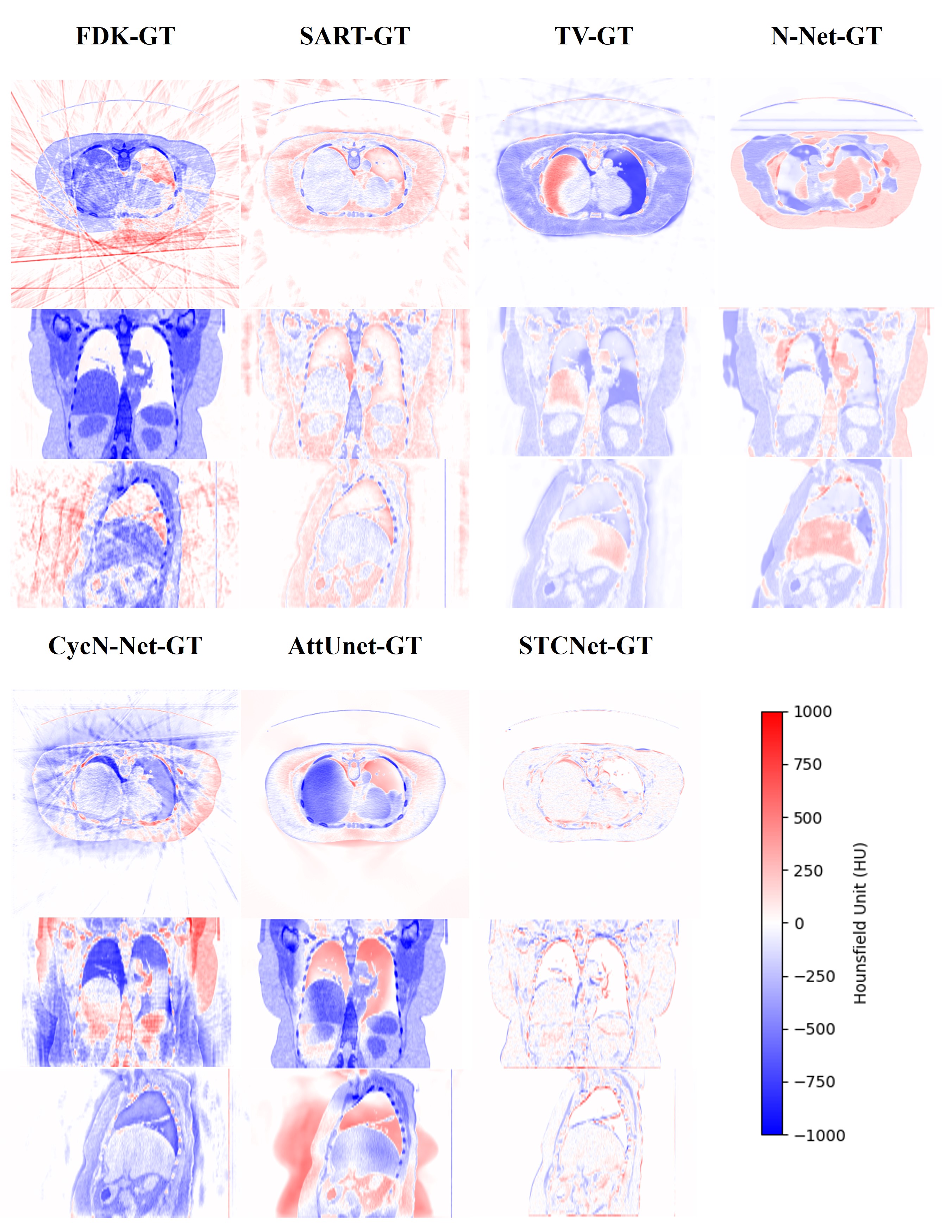}
  \caption{HU difference maps on the TCIA-LUNG dataset: The maps show axial, coronal, and sagittal differences between reconstructed images and ground truth. Red and blue represent positive and negative HU errors.}
  \label{fig:diff_TCIA}
\end{figure}

We also compared HU difference maps on the TCIA-LUNG dataset (Figure \ref{fig:diff_TCIA}). These maps show the difference between each reconstruction and the ground truth.
The FDK method showed large error regions across all planes. The largest errors appeared near dense structures and tissue boundaries. SART reduced part of the error, but it still showed scattered noise-like deviations. SART-TV reduced some noisy errors, but it also produced boundary loss in transition areas.

N-Net and CycN-Net still showed clear deviations near the lung boundary and mediastinal regions. AttUNet reduced some errors, but noticeable distortion remained in difficult areas.
PDGM reduced HU errors by a large margin. This result shows that projection completion is important for sparse-view reconstruction. STCNet reduced the errors even more after adding image-domain correction. The remaining errors became smaller and more localized. This pattern was clear near the pulmonary hilum, airway regions, and soft tissue interfaces.

\subsection{Ablation Study}
\label{ssec:ablation}

\begin{figure}[!h]
\centerline{\includegraphics[width=\columnwidth]{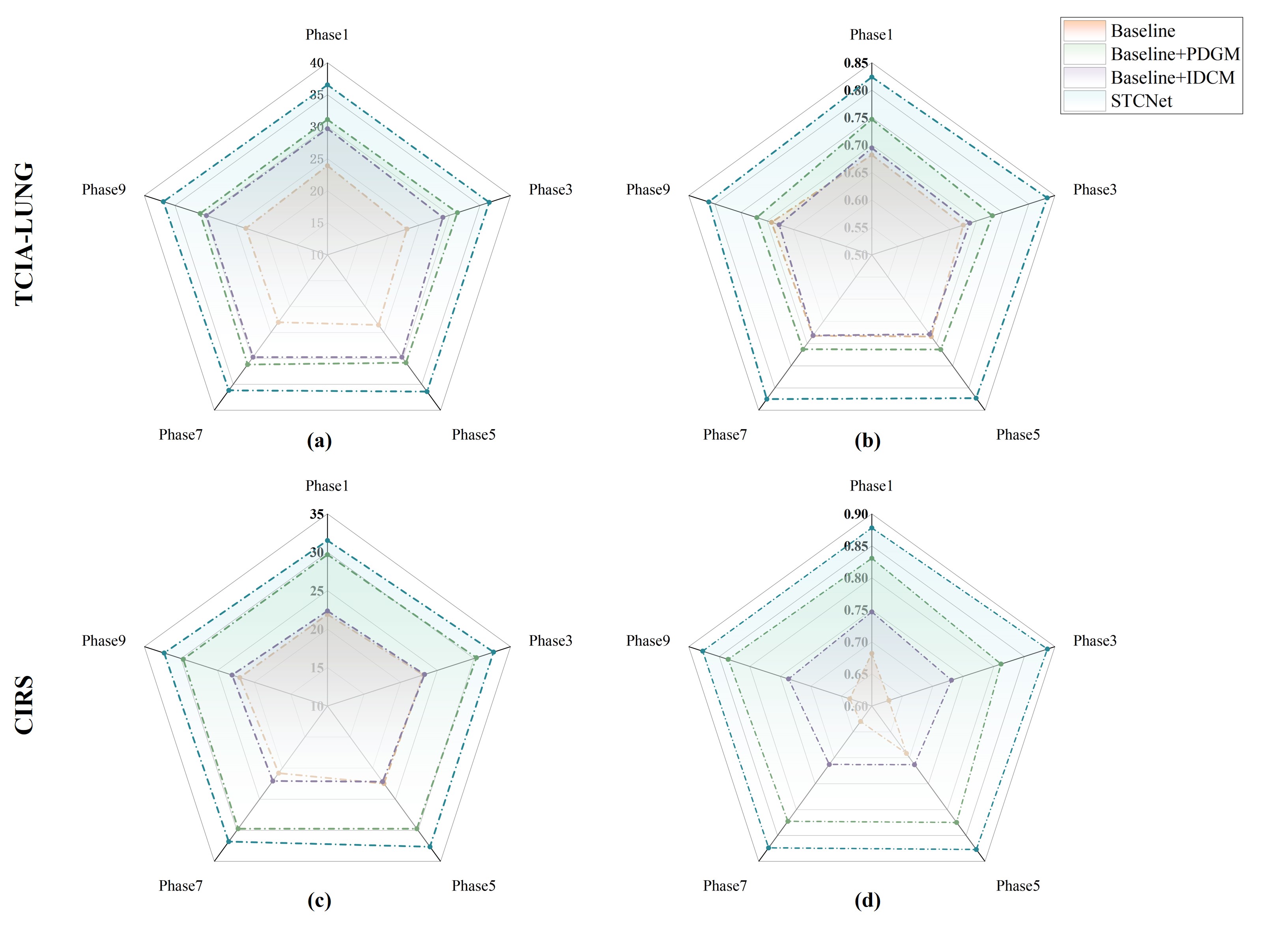}}
\caption{Ablation study of STCNet: The figure shows phase-wise PSNR and SSIM results for the baseline, baseline + PDGM, baseline + IDCM, and STCNet. Panels (a) and (b) show TCIA-LUNG results. Panels (c) and (d) show CIRS results.} 
\label{fig:ablation}
\end{figure}

We used ablation experiments to test the role of each module. Tables \ref{tab:ablation_tcia}, \ref{tab:ablation_cirs}, and Figure \ref{fig:ablation} show the results.

\begin{table}[!h]
  \centering
  \caption{Ablation study of PDGM and IDCM on the TCIA-LUNG dataset}
  \label{tab:ablation_tcia}
  \footnotesize
  \begin{tabular}{cccccc}
    \midrule
    \toprule[1pt]
    Baseline & PDGM & IDCM & MAE$\downarrow$ & PSNR$\uparrow$ & SSIM$\uparrow$ \\
    \midrule
    $\surd$ & & & 0.162 & 23.916 & 0.382 \\
    $\surd$ & $\surd$ & & \underline{0.059} & \underline{30.352} & \underline{0.747} \\
    $\surd$ & & $\surd$ & 0.162 & 22.360 & 0.695 \\
    $\surd$ & $\surd$ & $\surd$ & \textbf{0.036} & \textbf{32.536} & \textbf{0.824} \\
      \midrule
    \bottomrule[1pt]
  \end{tabular}
  \begin{tablenotes}
    \footnotesize
    \item[*] Remark: the PDGM and IDCM represent Projection Domain Generation Module and Image Domain Correction Module. The \textbf{Bold} and \underline{underline} indicate this metric's best and second-best performance, respectively.
  \end{tablenotes}
\end{table}

\begin{table}[!h]
  \centering
  \caption{Ablation study of PDGM and IDCM on the CIRS phantom dataset}
  \label{tab:ablation_cirs}
  \footnotesize
  \begin{tabular}{cccccc}
    \midrule
    \toprule[1pt]
    Baseline & PDGM & IDCM & MAE$\downarrow$ & PSNR$\uparrow$ & SSIM$\uparrow$ \\
    \midrule
    $\surd$ & & & 0.102 & 24.916 & 0.332 \\
    $\surd$ & $\surd$ & & \underline{0.064} & \underline{31.134} & \underline{0.783} \\
    $\surd$ & & $\surd$ & 0.162 & 29.697 & 0.695 \\
    $\surd$ & $\surd$ & $\surd$ & \textbf{0.018} & \textbf{36.543} & \textbf{0.878} \\
      \midrule
    \bottomrule[1pt]
  \end{tabular}
  \begin{tablenotes}
    \footnotesize
    \item[*] Remark: the PDGM and IDCM represent Projection Domain Generation Module and Image Domain Correction Module. The \textbf{Bold} and \underline{underline} indicate this metric's best and second-best performance, respectively.
  \end{tablenotes}
\end{table}

On the TCIA-LUNG dataset, the baseline model reached an MAE of 0.162, a PSNR of 23.916 dB, and an SSIM of 0.382. When we added PDGM, the model substantially improved. The MAE dropped to 0.059. The PSNR increased to 30.352 dB. The SSIM increased to 0.747. These changes show that projection-domain generation is a key part of the framework.

When we added IDCM alone, the model also improved. But the gain was smaller than the gain from PDGM. This result suggests that image-domain correction is useful, but it depends on a good initial reconstruction.
When we combined PDGM and IDCM, the model achieved the best results on both datasets. On TCIA-LUNG, STCNet reached an MAE of 0.036, a PSNR of 32.536 dB, and an SSIM of 0.824. On CIRS, STCNet reached an MAE of 0.018, a PSNR of 36.543 dB, and an SSIM of 0.878. Figure 6 shows that this advantage remained stable across phases. These findings support the idea that the two modules work well together.

\section{Discussion}
\label{sec:discussion}
This study presents STCNet, a dual-domain framework for sparse-view 4D CBCT reconstruction. The method combines projection completion and image refinement. The method also models both spatial geometry and respiratory motion. The results show that STCNet improves image quality on both TCIA-LUNG and CIRS data.

The results support two main points. First, projection-domain recovery matters a lot in sparse-view 4D CBCT. PDGM alone already produced large gains over the baseline and other methods. This result suggests that early recovery of missing projection information is important. Second, image-domain correction adds further value after reconstruction. IDCM reduced residual artifacts and improved local structure. The best performance came from combining both modules.

The spatial-temporal design also played an important role. Positional encoding helped the model learn the geometric relation between sparse angles. Temporal encoding helped the model use the continuity between breathing phases. These two parts likely explain why the method preserved phase-wise structure better than the comparison methods.

The study also has practical value. Sparse-view reconstruction can help reduce imaging dose and scan burden in 4D CBCT. STCNet used only 16 projections per phase and still produced strong results. This suggests that the method may support lower-dose imaging protocols in radiotherapy guidance. Improved phase-resolved image fidelity may also benefit downstream radiotherapy tasks, such as target localization, motion assessment, and adaptive decision-making, although these applications were not directly evaluated here.

Still, this work has several limitations. The first limitation is the sampling setting. We mainly trained and tested the model under a 16-view-per-phase design. The current results do not show how the model performs under more extreme sparse-view settings, such as 10 views or fewer. The second limitation is the lack of explicit anatomical priors. The method does not directly use structural knowledge about airways, vessels, or organ topology. This may reduce reconstruction reliability in cases with larger motion or more complex anatomy. The third limitation is the evaluation scope. We focused on image reconstruction metrics and visual quality. We did not test downstream tasks such as tumor localization, motion estimation, or dose calculation.

\textbf{Future work} can address these points. One direction is to add anatomical priors from 4D CT or MRI. Another direction is to design adaptive view selection strategies for sparse acquisition. A third direction is to test the method in task-based radiotherapy settings. These steps would help clarify the clinical value of the framework.

\section{Methodology}
\label{sec:methodology}
\begin{figure*}[!h]
\centering
\includegraphics[width=5in]{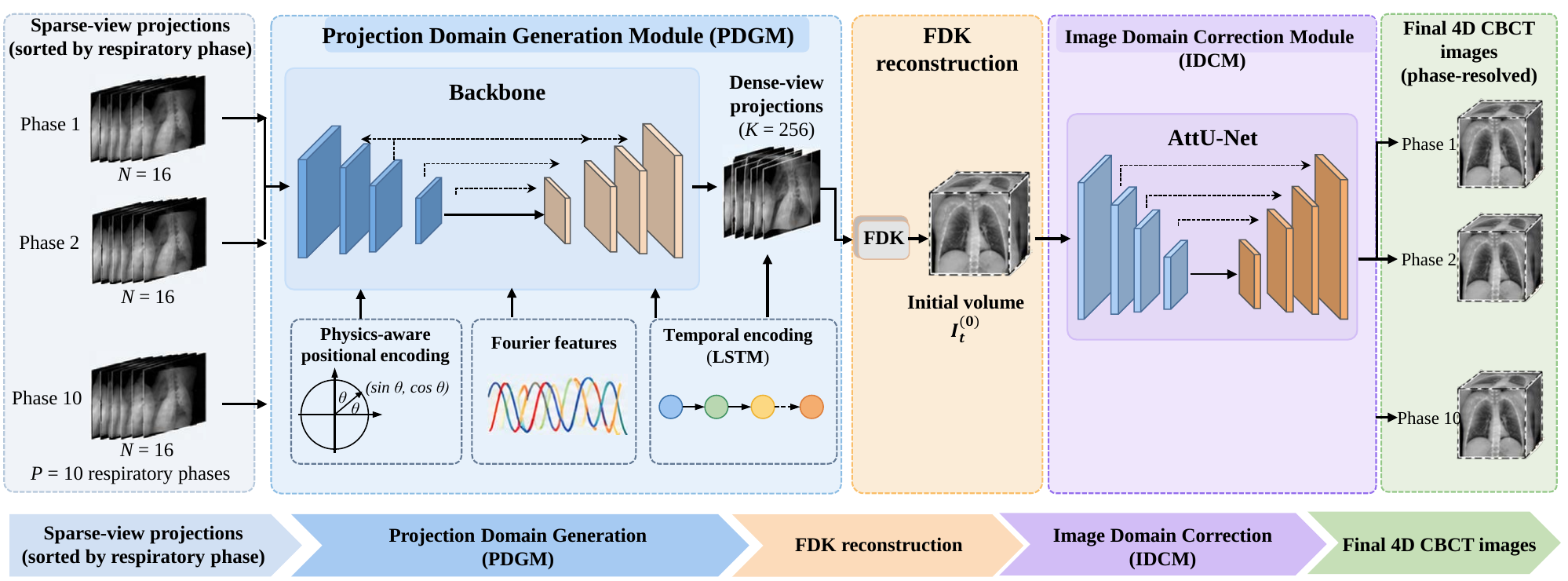}
\caption{ Overview of the STCNet framework: The workflow starts with sparse-view projections sorted by respiratory phase. The PDGM uses a Swin-UNet backbone with positional and temporal encoding to generate dense-view projections. The method then applies FDK reconstruction. The IDCM refines the reconstructed volumes and outputs final phase-resolved 4D CBCT images.}
\label{workflow}
\end{figure*}

In this section, We will introduce STCNet in detail, including Problem Definition, Projection Domain Generation Module (PDGM), Image Domain Correction Module (IDCM), Total Loss and some details of the network architecture. The workflow is shown in Figure \ref{workflow}.

\subsection{Architecture Overview}
\label{ssec:Architecture Overview}

We formulate sparse-view 4D CBCT reconstruction as a phase-resolved volumetric imaging problem with
\(P=10\) respiratory phases. For each phase \(t\), a sparse projection set is denoted as
\begin{equation}
\label{eq:S_t}
\mathcal{S}_t \;=\; \left\{ p_n^{(t)} \right\}_{n=1}^{N},
\end{equation}
where \(N=16\) is the number of sparse projections and \(p_n^{(t)}\) is the projection acquired at angle
\(\theta_n^{(t)}\). The corresponding angle set is
\begin{equation}
\label{eq:Theta_t}
\boldsymbol{\Theta}_t \;=\; \left\{ \theta_n^{(t)} \right\}_{n=1}^{N},
\end{equation}

Given the sparse-view input \(\mathcal{S}_t\), the goal is to reconstruct a high-quality phase-resolved image volume
\(\mathcal{I}_t\) for each respiratory phase.

The PDGM first maps the sparse-view projection set to a dense-view projection representation:
\begin{equation}
\label{eq:P_hat}
\hat{\mathcal{P}}_t \;=\; F_{\mathrm{PDGM}}(\mathcal{S}_t,\boldsymbol{\Theta}_t)
\end{equation}
where
\begin{equation}
\label{eq:Phat_k}
\hat{\mathcal{P}}_t \;=\; \left\{ \hat{p}_k^{(t)} \right\}_{k=1}^{K},
\end{equation}
and \(K=256\) is the number of synthesized projections for each phase.

The densified projections are then reconstructed by FDK to obtain the initial image-domain volume:
\begin{equation}
\label{eq:I_t0}
\mathcal{I}_t^{(0)} \;=\; \mathrm{FDK}(\hat{\mathcal{P}}_t), \qquad t=1,\dots,P \qquad
\end{equation}

The image-domain correction module then refines the initial reconstruction:
\begin{equation}
\label{eq:I_t}
\mathcal{I}_t \;=\; F_{\mathrm{IDCM}}(\mathcal{I}_t^{(0)},\, t=1,\dots,P)
\end{equation}

Accordingly, the overall STCNet pipeline can be expressed as
\begin{equation}
\label{eq:STCNet_def}
F_{\text{STCNet}}=F_{\text{IDCM}} \circ F_{\text{FDK}} \circ F_{\text{PDGM}}
\end{equation}

\subsection{Projection Domain Generation Module (PDGM)}
\label{ssec:PDGM}

The PDGM is designed to generate dense-view projections from sparse-view input within each respiratory phase. This step addresses the intrinsic data insufficiency of sparse-view 4D CBCT and provides a more complete projection set for subsequent reconstruction. During training, the generated dense-view projections are supervised by the corresponding ground-truth projection set \( \hat{\mathcal{P}}^{gt}_t \). The optimization target is
\begin{equation}
\label{eq:PDGM_obj}
\underset{(\hat{\mathcal{P}}_t,\hat{\mathcal{P}}^{gt}_t) \in \mathcal{D}}{\arg\min} \; \left\| \hat{\mathcal{P}}^{gt}_t - \hat{\mathcal{P}}_t \right\|_2^2 ,
\end{equation}
where \(\mathcal{D}\) denotes the projection-domain training dataset.

We used Swin-UNet as the backbone architecture of the PDGM.[23] For a sparse projection tensor \(S_t \in \mathbb{R}^{c\times n\times h\times w}\), the encoder extracts hierarchical projection features:
\begin{equation}
\label{eq:Xs}
X_s = \text{Encoder}(S_t),
\end{equation}
where \(X_s \in \mathbb{R}^{c'\times n'\times h'\times w'}\) denotes the latent feature representation.

\textbf{Physics-aware positional encoding}
Because sparse-view reconstruction is strongly constrained by acquisition geometry, we introduced a physics-aware positional encoding strategy to preserve angular relationships among projection views. Instead of using a simple normalized angle index, each acquisition angle \(\theta\) is mapped to the unit circle as
\begin{equation}
\label{eq:v_theta}
v(\theta) = (\sin \theta, \cos \theta),
\end{equation}
which explicitly preserves the circular geometry of rotational acquisition and more faithfully represents the relative relationship between projection angles.

To improve the representation of fine geometric variation, we further applied Fourier feature mapping to the positional representation. This design was motivated by the spectral bias of neural networks, which tend to learn low-frequency functions more easily than high-frequency ones.[33] For an input coordinate \(p\), the positional encoding is defined as
\begin{equation}
\label{eq:gamma}
\gamma(p) = \left( \sin(2^0 \pi p), \cos(2^0 \pi p), \dots, \sin(2^{L-1} \pi p), \cos(2^{L-1} \pi p) \right),
\end{equation}
where in our implementation, we set \(L=10\) for spatial coordinates and \(L=4\) for angular coordinates to balance representation capacity and computational cost.

\textbf{Temporal consistency encoding}
A key property of 4D CBCT is that anatomical structures evolve continuously across respiratory phases rather than changing independently from one phase to another. To exploit this temporal continuity, we introduced a temporal encoding mechanism in the latent feature space. After Swin-UNet encoding, the feature maps are reshaped into sequential embeddings and arranged in respiratory phase order. These phase-wise latent features are then processed by a long short-term memory (LSTM) module to model inter-phase dependencies and breathing dynamics.

Compared with directly applying recurrent operations to raw high-resolution projections, temporal modeling in the latent feature space is more efficient and more stable. It allows the network to capture phase-specific motion patterns while preserving compact semantic information. Let the temporal embedding be denoted by \(H_t\). The spatial feature, positional encoding, and temporal feature are fused as
\begin{equation}
\label{eq:X_fused}
X = X_s + \text{PE} + H_t,
\end{equation}
where PE is the positional encoding term. The fused feature \(X\) is then passed through the decoder to generate dense-view projections:
\begin{equation}
\label{eq:P_hat_t}
\hat{\mathcal{P}}_t = \text{Decoder}(X),
\end{equation}

After projection generation, FDK reconstruction is applied to obtain the initial volumetric image for each phase:
\begin{equation}
\label{eq:I_t0}
I_t^{(0)} = \text{FDK}(\hat{\mathcal{P}}_t),
\qquad t = 1, \dots, P.
\end{equation}


\subsection{Image domain correction module}
Although projection completion substantially improves sparse-view reconstruction, residual artifacts and local distortions may still remain after FDK reconstruction. We therefore introduced an image-domain correction stage to further refine the reconstructed volumes.

The IDC M takes the initial reconstructed image
\[
I_t^{(0)} \in \mathbb{R}^{c \times h \times w}
\]
as input and produces the corrected image \(I_t\). We implemented this module using an AttU-Net-based architecture. During training, the image-domain refinement is supervised by the corresponding reference image \(I_t^{gt}\), with the optimization target
\begin{equation}
\label{eq:IDC_obj}
\underset{(\,\hat{I}_t,\, I_t^{gt}\,)\in \mathcal{D}_1}{\arg\min} \; \left\| I_t^{gt} - I_t \right\|_2^2 .
\end{equation}
where \(\mathcal{D}_1\) denotes the image-domain training dataset.

The corrected output is defined as
\begin{equation}
\label{eq:I_t}
I_t = \text{IDC M}(I_t^{(0)}).
\end{equation}

\subsection{Loss Function}

\textbf{Projection-domain loss}:
The PDGM was trained using a composite loss that combines mean squared error (MSE), structural similarity (SSIM) loss, and adjacent-slice gradient consistency loss.

The MSE term is defined as
\begin{equation}
\label{eq:L_MSE}
L_{\text{MSE}} = \frac{1}{N_p} \sum_{i=1}^{N_p} \left( \hat{p}_{gt}(i) - \hat{p}_t(i) \right)^2,
\end{equation}
where \(N_p\) denotes the total number of pixels in the projection set.

To preserve structural similarity between generated and reference projections, we used an SSIM-based loss
\begin{equation}
\label{eq:L_SSIM}
L_{\text{SSIM}} = 1 - \text{SSIM}(\hat{p}_t, \hat{p}_{gt}).
\end{equation}

To encourage local continuity and reduce structural inconsistency between adjacent slices, we further introduced a gradient consistency loss
\begin{equation}
\label{eq:L_S}
L_{S} = \frac{1}{N_p} \sum_{i=1}^{N_p} \left\| \nabla \hat{p}_t(i, z) - \nabla \hat{p}_t(i, z+1) \right\|_2^2,
\end{equation}
where \(\hat{p}_t(i,z)\) and \(\hat{p}_t(i,z+1)\) denote adjacent slices along the depth direction, and \(\nabla\) represents the spatial gradient operator.

The final PDGM training objective is
\begin{equation}
\label{eq:L_total_PD}
L_{\text{PD}} = \alpha L_{\text{MSE}} + \beta L_{\text{SSIM}} + \lambda L_S,
\end{equation}
with empirically set constants \(\alpha, \beta, \lambda\).

\textbf{Image-domain loss}:
The IDC M is trained using a combined image-domain loss composed of MSE and SSIM terms. The MSE component is
\begin{equation}
\label{eq:L_MSE_img}
L'_{\text{MSE}} = \frac{1}{M} \sum_{j=1}^{M} \left( I_{\text{gt}}(j) - I(j) \right)^2,
\end{equation}
where \(M\) denotes the number of image pixels or voxels.

The image-domain SSIM loss is
\begin{equation}
\label{eq:L_SSIM_img}
L'_{\text{SSIM}} = 1 - \text{SSIM}(I, I_{\text{gt}}).
\end{equation}

The final image-domain training objective is
\begin{equation}
\label{eq:L_total_img}
L'_{\text{total}} = \delta L'_{\text{MSE}} + \epsilon L'_{\text{SSIM}},
\end{equation}
and in all experiments we set \(\delta = 0.4\) and \(\epsilon = 0.6\).

\subsection{Reconstruction procedure}

For each respiratory phase, STCNet reconstructs the final image in three steps. First, the sparse-view projection subset \(S_t\) and its known acquisition angles \(\theta_t\) are used as input to the PDGM. Second, the PDGM generates a dense-view projection set \(\hat{\mathcal{P}}_t\), which is reconstructed by FDK to obtain the initial image volume \(I_t^{(0)}\).
Third, the IDCM refines \(I_t^{(0)}\) to produce the final corrected image \(I_t\). Repeating this process for all 10 respiratory phases yields the full 4D CBCT sequence.

\section{Declarations}\label{sec:Declarations}

\noindent \textbf{Data Availability}

The TCIA-LUNG dataset is publicly available through The Cancer Imaging Archive \cite{r15}. The associated NSCLC 4D lung imaging data used in this study are described in \cite{r16}. The CIRS phantom data are available from the corresponding author upon reasonable request and subject to institutional policy.

\noindent \textbf{Code availability}

Code has publicly available at: \url{https://github.com/Lai-KK/STCNet.git}.


\noindent \textbf{Author contribution}

QI LAI: Conceptualization, Methodology, Formal analysis, Writing-Original draft; 
YUTONG HE: Methodology, Validation, Experiments;
ZHICHENG LI: Data curation, Investigation, Resources;
LEI WANG: Writing Review,Funding acquisition;
YAN YAN: Software,Validation,Data curation;
YANZHONG MA: Investigation, Data curation;
XIAOKUN LIANG: Writing Review, Funding acquisition;
DONG LIANG: Data curation, Investigation, Resources;
LEI JIN: Writing Review, Funding acquisition.

\noindent \textbf{Conflict of interest/Competing interests}

The authors declare that they have no conflicts of interest or competing interests related to this work.

\noindent \textbf{Acknowledgments}

This work was supported by the National Key Research and Development Program of China (2024YFF1206901); in part by the Shenzhen General Program under Grant Ref. JCYJ20240813154703005; in part by the National Natural Science Fund for Young Scientists Fund Program C (82502447).

\bibliography{sn-bibliography}

\end{document}